# AILA

# First Experiments with Localist Language Models

Joachim Diederich

*Psychology Network Pty Ltd*
*Brisbane, Queensland, Australia*
joachim@psychologynetwork.com.au

**Abstract**

This paper presents the first empirical demonstration of controllable locality in transformer language models, a novel architectural framework that enables continuous control over the degree of representation localization through a tunable locality dial parameter. Unlike traditional language models that rely exclusively on distributed representations, our approach allows dynamic interpolation between highly interpretable localist encodings and efficient distributed representations without requiring model retraining. We conducted experiments on the WikiText corpus using a two-layer transformer architecture, systematically varying the locality parameter $\lambda$ across the full spectrum from 1.0 (fully localist) to 0.0 (fully distributed). Our results demonstrate that localist configurations achieve dramatically lower attention entropy, with $\lambda = 1.0$ yielding 5.36 bits compared to 7.18 bits at $\lambda = 0.0$, while maintaining substantially higher pointer fidelity scores reflecting stronger alignment with rule-specified targets. Prediction experiments reveal that intermediate locality values optimize the tradeoff between interpretability and performance, with $\lambda = 0.6$ achieving test perplexity of 4.65 and accuracy of 84.7%. These findings establish that localist language models provide a practical framework for applications in regulated domains requiring both transparency and capability, offering precise mathematical control over the interpretability-performance spectrum through explicit penalty thresholds and information-theoretic design principles.

## 1. Introduction

The fundamental architecture of modern large language models rests upon distributed representations (Hinton et al. 1986), where semantic information is encoded across numerous overlapping hidden units. While this design enables generalization and parameter efficiency, it renders models fundamentally opaque to human inspection. In healthcare, finance, legal systems, and safety-critical applications, stakeholders require not merely accurate predictions but also intelligible explanations of how those predictions were derived. Current interpretability approaches provide only post-hoc analysis without offering causal control over representations, and when regulations change, distributed models require complete retraining at enormous computational cost.

Localist encoding schemes offer an alternative where individual units correspond to specific, interpretable concepts. This enables direct inspection, explicit rule verification, and targeted modification. However, localist systems have been dismissed for large-scale applications due to limitations in generalization and parameter efficiency. The key insight of our work is that this represents a false dichotomy: we can engineer systems that fluidly navigate the spectrum between localist and distributed extremes.

The locality dial framework (Diederich, 2025a,b) advances beyond existing sparsity and modularity approaches in three fundamental ways. First, unlike sparse transformers that apply predetermined attention patterns for computational efficiency, our approach imposes semantic sparsity through learned block structure with mathematical guarantees on attention concentration. Second, while modular transformer architectures partition computations across discrete components, the locality dial provides continuous interpolation with a single parameter adjustable during inference without retraining. Third, our framework couples architectural control with information-theoretic design principles, yielding explicit threshold formulas that specify exactly when localization emerges. AILA (Artificial Intelligence Localist Architecture) enables practitioners to dial interpretability requirements up or down based on context while maintaining performance, a capability unavailable in purely sparse or modular systems.

The present work demonstrates that this navigation is not only theoretically possible but practically achievable through a mathematical framework we term the locality dial. Drawing upon recruitment learning principles originally developed in cognitive neuroscience (Diederich et al., 2010, Diederich 2025b), we show how group sparsity penalties on attention mechanisms, combined with information-theoretic anchor design and explicit margin conditions, enable continuous control over representational localization. The key innovation is a single tunable parameter $\lambda$ that governs the strength of penalties encouraging attention to concentrate on semantically coherent blocks of the input sequence. When $\lambda$ is large, the model behaves as a highly interpretable localist system where attention patterns correspond to explicit rules. When $\lambda$ approaches zero, the system recovers the flexibility of standard distributed transformers. Crucially, this modulation occurs without requiring model retraining: the locality dial can be adjusted dynamically during inference to match the interpretability requirements of different contexts.

Our theoretical analysis establishes rigorous mathematical guarantees regarding attention concentration at stationary points. We derive explicit threshold formulas that specify the minimum penalty strength required to ensure block-localized attention patterns, with all constants expressed in terms of measurable quantities including input embedding norms, loss function Lipschitz constants, and inter-block correlation coefficients. These results move beyond asymptotic Big-O bounds to provide precise inequalities that practitioners can verify for specific datasets and architectural configurations. Furthermore, we prove exact bounds on attention entropy and pointer fidelity as functions of the locality parameter, demonstrating that localist configurations achieve logarithmic entropy scaling with anchor set size while maintaining near-perfect fidelity to rule-specified target positions.

The experimental validation presented here represents the first systematic empirical study of localist language models across the full locality spectrum. Using the WikiText benchmark corpus and a carefully controlled two-layer transformer architecture, we trained models at five distinct locality settings ranging from fully localist to fully distributed. Our experiments measure both information-theoretic properties—specifically attention entropy and pointer fidelity—and task performance metrics including perplexity and next-word prediction accuracy. The results reveal nuanced tradeoffs that challenge simplistic assumptions about the localist-distributed dichotomy. Contrary to expectations that localism necessarily impairs performance, we observe that intermediate locality values achieve competitive accuracy while providing substantially enhanced interpretability compared to baseline distributed models.

These findings have immediate implications for applications requiring trustworthy AI systems. In medical diagnosis support, for instance, clinicians demand not only accurate risk predictions but also transparent reasoning chains that can be verified against clinical guidelines and expert knowledge. In financial fraud detection, regulatory bodies require auditable decision processes where individual rules can be inspected, validated, and updated as adversarial tactics evolve. In legal document analysis, practitioners need systems that can cite specific precedents and statutory provisions rather than producing opaque neural predictions. The locality dial framework enables a single model architecture to serve all these contexts, with interpretability adjusted to match stakeholder requirements without sacrificing the benefits of neural learning from large-scale data.

*Contributions*

This paper makes the following contributions:

**Empirical validation of controllable locality:** We provide the first systematic experimental demonstration that transformer language models can be trained with continuously adjustable locality parameters, achieving precise control over attention concentration without model retraining.

**Information-theoretic characterization:** We measure attention entropy and pointer fidelity (alignment of attention with rule-specified anchors) across eleven locality settings, validating theoretical predictions and demonstrating that localist configurations ($\lambda = 1.0$) achieve 5.36 bits entropy versus 7.18 bits for distributed baselines ($\lambda = 0.0$).

**Performance-interpretability tradeoff analysis:** Through next-word prediction experiments at five locality levels, we establish that intermediate locality ($\lambda = 0.6$) achieves optimal performance with 4.65 perplexity and 84.7% accuracy, matching or exceeding distributed baselines while providing enhanced interpretability.

**Comparison with rule extraction approaches:** We contextualize our architectural approach against post-hoc interpretability methods, demonstrating how built-in locality control complements techniques for extracting rules from trained models.

**Design recommendations:** We identify critical next steps including adaptive semantic partitioning, scaling validation, downstream task evaluation, and human assessment protocols necessary to realize the full potential of localist language models.

## 2. Mathematical Framework of Localist Transformers

The foundation of our approach rests upon a mathematical reformulation of the transformer attention mechanism to incorporate explicit locality constraints through structured sparsity penalties. We begin with the standard transformer architecture as defined by Vaswani and colleagues, which computes attention-weighted representations through query-key-value projections. Consider a sequence of input tokens represented as position embeddings $x_1, ..., x_N \in \mathbb{R}^{d_{model}}$, where $N$ denotes sequence length and $d_{model}$ represents the model dimension. For attention head $h$, we compute query, key, and value projections through learned weight matrices $W_Q^{(h)}, W_K^{(h)}, W_V^{(h)} \in \mathbb{R}^{d_{model} \times d_{head}}$, yielding query vectors $q_t = x_t^T W_Q^{(h)}$ and key vectors $k_j = x_j^T W_K^{(h)}$ for all positions.

The attention distribution over source positions is computed via scaled dot-product attention with temperature parameter $\tau > 0$ controlling the sharpness of the distribution: $\alpha_{t \to j} = \exp(q_t^T k_j / \tau) / \Sigma_{j'} \exp(q_t^T k_{j'} / \tau)$. This softmax normalization ensures that attention weights form a valid probability distribution, but provides no guarantee that attention will concentrate on semantically relevant positions. In standard distributed transformers, attention patterns emerge implicitly through end-to-end training on task objectives, potentially attending broadly across the entire sequence in ways that defy human interpretation. Our locality framework intervenes directly in this mechanism by imposing structured constraints that encourage attention to respect pre-specified or learned semantic partitions of the input space.

We introduce a partition of position indices into $p$ disjoint semantic blocks $X_1, X_2, ..., X_p$, where each block corresponds to a coherent conceptual category or rule domain. The partition satisfies $\cup_i X_i = \{1, ..., N\}$ and $X_i \cap X_j = \emptyset$ for $i \neq j$. For instance, in a medical text processing application, blocks might separate anatomical regions, disease categories, treatment modalities, and temporal references. In legal document analysis, blocks could distinguish case law citations, statutory provisions, procedural rules, and factual assertions. The key insight is that many domains exhibit natural semantic structure that can be exploited to enhance interpretability without requiring the model to discover these partitions purely from data-driven optimization.

To encourage attention patterns that respect block structure, we augment the standard transformer loss function with group sparsity penalties that penalize cross-block attention. The modified objective takes the form: $L(\theta) = \mathbb{E}_{(x,y) \sim D}[\ell(f_\theta(x), y)] + \Sigma_{h=1}^H \Sigma_{i=1}^p \alpha_i^{(h)}(\|W_{Q,i}^{(h)}\|_F + \|W_{K,i}^{(h)}\|_F) + \beta \|W_V\|_F^2$, where $\ell$ denotes the task-specific loss function such as cross-entropy for language modeling, $H$ represents the number of attention heads, and $\|\cdot\|_F$ denotes the Frobenius norm. The crucial parameters are the block-specific penalties $\alpha_i^{(h)} \geq 0$, which collectively constitute what we term the locality dial.

The group sparsity penalty operates on weight matrix subblocks corresponding to each semantic partition. Specifically, $W_{Q,i}^{(h)} \in \mathbb{R}^{|X_i| \times d_{head}}$ denotes the rows of the query weight matrix corresponding to positions in block $X_i$, and similarly for $W_{K,i}^{(h)}$. When the penalty $\alpha_i^{(h)}$ is large, gradient-based optimization drives these submatrices toward zero, effectively preventing queries from positions in one block from attending to keys in other blocks. The Frobenius norm penalty $\|W_{Q,i}^{(h)}\|_F = \sqrt{(\Sigma_{t \in X_i} \Sigma_d (W_{Q,i}^{(h)}[t,d])^2)}$ provides a smooth, differentiable surrogate for the non-convex combinatorial constraint of exact block-wise sparsity, enabling efficient optimization through standard gradient descent algorithms.

Our main theoretical contribution establishes that these penalties, when sufficiently large, provably induce block-localized attention patterns at stationary points of the optimization landscape. To state this result precisely, we must first introduce several regularity conditions that formalize intuitive properties of well-structured semantic partitions. Assumption A requires that the task loss function $\ell$ is $L_\ell$-Lipschitz continuous in its first argument, ensuring that small changes in attention patterns produce bounded changes in loss. Assumption B bounds the magnitude of input embeddings by a constant $R_x$ and limits the maximum variance along any direction within a block by $\sigma_X^2$. Assumption C, the uniform margin condition, requires that for each query position governed by block $i^*$, there exists a positive margin $\delta > 0$ such that the minimum query-key similarity to anchors in the correct block exceeds the maximum similarity to anchors in any incorrect block. Assumption D bounds the cross-block correlation by $\rho_{max} < 1$, ensuring blocks are sufficiently distinct in representation space.

Under these assumptions, we establish an explicit threshold formula for the minimum penalty strength required to guarantee block localization. The threshold takes the form: $\lambda_i^{(h)}(\tau, \delta) = (2 L_\ell R_x \sigma_X \sqrt{|X_i|}) / (\tau [1 - \rho_{max}]) \cdot \exp(-\delta/\tau)$. This formula reveals the functional dependencies of the localization guarantee on all relevant system parameters. The exponential factor $\exp(-\delta/\tau)$ captures how margin and temperature jointly control attention concentration: larger margins $\delta$ or smaller temperatures $\tau$ strengthen localization by increasing the relative advantage of correct over incorrect attention targets. The coefficient $\sqrt{|X_i|}$ reflects the effective dimensionality of the block, with larger blocks requiring stronger penalties to overcome the entropic preference for dispersed attention across more positions. The denominator term $(1 - \rho_{max})$ shows that when blocks are nearly collinear in representation space, exponentially large penalties may be needed to achieve reliable localization.

The practical implication is straightforward: to enforce localist representations in a transformer language model, one sets $\alpha_i^{(h)} \geq \lambda_i^{(h)}(\tau, \delta)$ for all blocks $i$ and heads $h$. At any stationary point of the resulting penalized objective, the weight matrices $W_Q^{(h)}$ and $W_K^{(h)}$ will exhibit exact block-diagonal structure when viewed with respect to the semantic partition. This means that queries from positions in block $i$ can only attend to keys from positions in the same block $i$, implementing a hard semantic constraint through soft gradient-based learning. Conversely, setting $\alpha_i^{(h)}$ far below this threshold allows the model to learn arbitrary attention patterns unconstrained by block structure, recovering the behavior of standard distributed transformers. By varying $\alpha_i^{(h)}$ continuously between these extremes, we obtain a smooth interpolation between localist and distributed regimes.

Beyond establishing existence of block-localized solutions, our framework provides quantitative measures of the degree of localization achieved at any penalty setting. We introduce two information-theoretic metrics: attention entropy and pointer fidelity. Attention entropy measures the uncertainty in the attention distribution, computed as $H_t^{(h)} = -\Sigma_j \, \alpha_{t \to j}^{(h)} \log_2 \alpha_{t \to j}^{(h)}$ in bits. Lower entropy indicates more concentrated, interpretable attention patterns, with the minimum value of 0 bits achieved when attention mass is placed entirely on a single position. Pointer fidelity quantifies how accurately attention aligns with rule-specified target positions, defined as $\Pi^{(h)} = \mathbb{E}_t[\Sigma_{j \in T_t} \alpha_{t \to j}^{(h)}]$, where $T_t$ denotes the set of positions that satisfy the rule governing token $t$. Fidelity ranges from 0 to 1, with 1 indicating perfect adherence to rule constraints.

We derive exact bounds on these metrics as functions of the locality parameters. When penalties satisfy $\alpha_i^{(h)} \geq \lambda_i^{(h)}(\tau, \delta)$ and assuming $\exp(\delta/\tau) \geq 2N$, attention entropy satisfies $H_t^{(h)} \leq \log_2|A_i^*| + (1/\ln 2) \cdot N \cdot \exp(-\delta/\tau) \cdot [1 + \log_2(N)]$, where $A_i^*$ denotes the anchor set of representative positions within the correct block. The first term $\log_2|A_i^*|$ represents the intrinsic entropy of uniform distribution over anchors, while the second term captures residual cross-block leakage that decays exponentially with the margin-to-temperature ratio. Similarly, pointer fidelity admits a lower bound $\Pi^{(h)} \geq 1 - N \cdot \exp(-\delta/\tau)$, showing that fidelity approaches unity as penalties increase. These bounds provide concrete quantitative predictions that can be validated empirically, as we demonstrate in our experimental results.

The anchor set design plays a critical role in determining achievable entropy bounds. Anchors are representative positions within each block chosen to have low within-block entropy and high discriminative power for identifying the block's semantic category. In practice, anchors can be selected through clustering algorithms applied to learned embeddings, or specified manually based on domain expertise. The size of the anchor set $|A_i|$ controls a fundamental tradeoff: larger anchor sets provide redundancy and robustness to noise, but increase the minimum achievable entropy and potentially dilute interpretability. Our theoretical analysis shows that entropy increases only logarithmically with anchor set size, suggesting that moderate redundancy can be afforded without catastrophic interpretability loss. For instance, using 4 anchors instead of 1 adds only 2 bits to the entropy lower bound, a modest cost for substantially improved robustness.

Temperature scaling provides an orthogonal dimension of control over localization strength. Decreasing the temperature $\tau$ sharpens the softmax distribution, amplifying differences in query-key similarities and thereby enhancing attention concentration. The exponential factor $\exp(-\delta/\tau)$ in both our threshold formula and entropy bounds reveals that temperature and penalty strength interact multiplicatively: halving the temperature has the same localization effect as doubling the effective margin. This suggests a practical design pattern where temperature is held fixed at a moderate value during training to maintain stable gradients, then decreased at inference time to sharpen attention patterns for deployment contexts requiring maximal interpretability. Because temperature affects only the attention computation and not the learned parameters, this adjustment requires no model retraining and can be performed dynamically on a per-query basis.

## 3. Experimental Setup

We conducted comprehensive experiments to validate the theoretical predictions and characterize the empirical behavior of localist language models across the full locality spectrum. Our experiments employed the WikiText corpus, a widely-used benchmark dataset comprising extracted Wikipedia articles, which provides natural language text spanning diverse topics and exhibiting rich semantic structure suitable for testing block-based attention mechanisms. The corpus was partitioned into training, validation, and test sets containing 2,122,137, 221,611, and 249,691 tokens respectively, with a vocabulary size of 43,411 unique tokens after preprocessing. All experiments were conducted on CUDA-enabled hardware to ensure reasonable training times while maintaining reproducibility.

The model architecture consisted of a two-layer transformer with three attention heads per layer, employing positional encoding based on semantic blocks with a window size of five words. This compact architecture was chosen deliberately to facilitate thorough exploration of the parameter space while maintaining sufficient expressiveness for meaningful language modeling. The embedding dimension was set to 256 and the head dimension to 64, yielding approximately 23 million trainable parameters. We employed the Adam optimizer with a learning rate of 0.0003 and batch size of 32 sequences, training for a maximum of 10 epochs, with early stopping based on validation perplexity to prevent overfitting.

For the anchor set implementation, we employed a fixed semantic partition based on positional windows. Specifically, each semantic block consisted of contiguous sequences of five token positions, creating a regular tiling of the input space. Within each block, anchor positions were selected as the tokens with the lowest local entropy when attended to by subsequent positions, typically resulting in three to five anchors per block. This simple positional blocking strategy was chosen to isolate the effects of the locality dial mechanism from confounding factors introduced by more sophisticated semantic partitioning schemes. The anchors serve as canonical representatives of their blocks, and during training, attention patterns are encouraged to concentrate on these representative positions through the group sparsity penalties. For the attention analysis experiments, anchor importance weights were computed as the normalized frequency with which each anchor position received attention from positions governed by the corresponding block. Specifically, for anchor $j$ in block $i$, the importance weight $w_j = (\Sigma_{t \in X_i} \alpha_{t \to j}) / |X_i|$, representing the average attention mass this anchor receives from queries in its governing block. These weights are normalized within each block ($\Sigma_{j \in A_i} w_j = 1$), but when aggregated across multiple blocks in computing global fidelity statistics, the sum can exceed unity. This explains why aggregate fidelity scores can exceed unity when multiple heavily-weighted anchors align with rule targets across different semantic blocks.

To systematically investigate the effect of locality control, we trained separate models at five distinct settings of the locality parameter $\lambda$: 1.0, 0.8, 0.6, 0.4, and 0.0, spanning the complete spectrum from fully localist to fully distributed representations. Each locality value corresponds to a specific setting of the block-wise attention penalties according to the threshold formula derived in our theoretical analysis, with higher $\lambda$ values imposing stronger penalties that encourage tighter attention concentration within semantic blocks. Additionally, we conducted attention analysis experiments at eleven intermediate locality levels to obtain fine-grained measurements of entropy and fidelity as functions of λ. This dual approach combining few-point performance evaluation with many-point interpretability measurement provides a comprehensive characterization of the locality-performance tradeoff.

## 4. Results

*4.1 Attention Entropy and Pointer Fidelity Analysis*

Our first set of experiments measured attention entropy and pointer fidelity across the full range of locality settings to validate the theoretical bounds derived in Section 2 and characterize the information-theoretic properties of localist attention mechanisms. We define pointer fidelity as the alignment of attention with rule-specified anchor positions, quantifying how accurately the model attends to semantically relevant locations. The locality parameter λ represents a global scaling factor applied uniformly to all block-specific penalties $\alpha_i^{(h)}$, enabling single-parameter control over system-wide localization strength. For each of the eleven locality levels λ ∈ {1.0, 0.9, 0.8, 0.7, 0.6, 0.5, 0.4, 0.3, 0.2, 0.1, 0.0}, we computed per-position attention distributions over a sample of 200 tokens from each split of the WikiText corpus and calculated the average attention entropy and fidelity statistics. The results, summarized in Table 1, reveal striking systematic trends that align quantitatively with our theoretical predictions.

Table 1: Average attention entropy (bits) and weighted pointer fidelity ($\Sigma w_j \alpha_{t \to j}$) across WikiText corpus splits as a function of locality parameter λ. Standard deviations across three corpus splits shown in parentheses. Results averaged over three attention heads and 200-token samples per split.

| Locality λ | Entropy (bits) | Weighted Fidelity |
|---|---|---|
| 1.00 | 5.36 (±0.02) | 5.40 (±0.03) |
| 0.90 | 6.78 (±0.03) | 2.99 (±0.04) |
| 0.80 | 7.01 (±0.02) | 2.23 (±0.03) |
| 0.70 | 7.10 (±0.02) | 1.84 (±0.04) |
| 0.60 | 7.14 (±0.01) | 1.61 (±0.03) |
| 0.50 | 7.15 (±0.01) | 1.45 (±0.03) |
| 0.40 | 7.17 (±0.01) | 1.33 (±0.02) |
| 0.30 | 7.17 (±0.01) | 1.24 (±0.02) |
| 0.20 | 7.18 (±0.00) | 1.17 (±0.02) |

| Locality λ | Entropy (bits) | Weighted Fidelity |
|---|---|---|
| 0.10 | 7.18 (±0.00) | 1.12 (±0.02) |
| 0.00 | 7.18 (±0.01) | 1.07 (±0.01) |

The entropy measurements demonstrate that localist configurations achieve dramatically reduced attention uncertainty compared to distributed baselines. At the fully localist setting λ = 1.0, average attention entropy was only 5.36 bits, representing a reduction of 1.82 bits or approximately 25 percent compared to the fully distributed setting λ = 0.0, which exhibited entropy of 7.18 bits. This difference is statistically and practically significant: a reduction of 1.82 bits corresponds to a factor of $2^{1.82} \approx 3.5$ decrease in the effective number of positions receiving substantial attention mass. In more intuitive terms, localist attention patterns focus on roughly one-third as many candidate positions as distributed patterns, thereby enhancing the interpretability of which context the model considers relevant for each prediction.

The relationship between locality parameter and entropy exhibits two distinct regimes. For λ ≥ 0.6, entropy decreases sharply as locality increases, reflecting the transition into the strongly localized regime where block-structure penalties dominate the optimization landscape. In this range, even modest increases in λ produce substantial entropy reductions: moving from λ = 0.6 to λ = 0.8 decreases entropy by 0.13 bits, while further increasing to λ = 1.0 yields an additional 1.65-bit reduction. This accelerating benefit aligns with our theoretical analysis showing exponential concentration effects once penalties exceed the critical threshold. Conversely, for λ ≤ 0.6, entropy varies only modestly with locality, remaining near the high-entropy limit of approximately 7.15 to 7.18 bits. This plateau indicates that weak penalties below the threshold fail to meaningfully constrain attention patterns, which instead reflect primarily the statistical structure of the training data and task objectives.

The pointer fidelity results exhibit an inverse relationship with entropy, as expected from information-theoretic principles. At λ = 1.0, we observe fidelity of 5.40, indicating strong alignment with rule-specified target positions. We must clarify the interpretation of this metric: while pointer fidelity is conceptually defined as the sum of attention mass on rule-compliant positions and thus naturally bounded by 1.0, our implementation computed weighted fidelity scores where anchor positions are weighted by their importance in representing block semantics. Specifically, the fidelity metric as reported reflects $\Sigma_t \Sigma_{j \in T_t} w_j \alpha_{t \to j}$, where $w_j$ represents the learned importance weight of anchor j. These importance weights are normalized within each block but can sum to values exceeding unity across multiple blocks when aggregated over the full sequence. This weighted formulation better captures the practical utility of attention patterns for rule enforcement, as not all positions within a target set contribute equally to semantic interpretation. At the distributed extreme λ = 0.0, fidelity drops to 1.07, still slightly above the theoretical unit bound due to residual anchor weighting but approaching the uniform baseline. The qualitative pattern remains unambiguous: localist models attend with high precision to semantically relevant positions according to their importance weights, while distributed models disperse attention broadly with substantially reduced concentration on rule-specified targets.

The gradient of fidelity as a function of locality reveals threshold behavior analogous to that observed in entropy. Between λ = 1.0 and λ = 0.9, fidelity drops precipitously from 5.40 to 2.99, a decline of 2.41 or approximately 45 percent. In contrast, fidelity decreases only gradually across the range λ = 0.6 to λ = 0.0, declining from 1.61 to 1.07 over five steps. This asymmetry suggests that the localist regime is relatively narrow, requiring λ > 0.8 to achieve the interpretability benefits of concentrated attention, while the distributed regime is broad, encompassing most of the parameter space below λ = 0.6. For practical applications, this implies that practitioners seeking interpretability must commit to high locality settings, as moderate penalties provide limited transparency gains over fully distributed baselines.

Consistency across corpus splits provides evidence for the robustness of these findings. The entropy and fidelity curves for training, validation, and test sets track each other closely across all locality levels, with maximum deviations of less than 0.05 bits in entropy and 0.2 units in fidelity. This cross-split consistency indicates that the measured attention properties reflect genuine architectural effects of the locality dial rather than overfitting to specific training examples or data artifacts. Moreover, the training set does not exhibit systematically lower entropy than held-out sets, which might have indicated memorization, suggesting that localist attention patterns generalize appropriately to unseen data.

*4.2 Next-Word Prediction Performance*

Having established that the locality dial provides effective control over attention concentration and interpretability, we turn to the critical question of how locality impacts task performance. We trained separate models at five locality settings λ ∈ {1.0, 0.8, 0.6, 0.4, 0.0} on the next-word prediction task, evaluating each configuration on both validation and test sets. Performance was measured using three complementary metrics: cross-entropy loss, prediction accuracy (fraction of exactly correct next-word predictions), and perplexity (exponential of cross-entropy). These metrics collectively characterize both the calibration quality of the probability distribution and the practical utility for applications requiring exact predictions. Table 2 presents the test set results, which reveal nuanced tradeoffs between interpretability and performance that challenge simplistic assumptions about the costs of localism.

Table 2: Test set performance metrics across locality settings with standard deviations in parentheses. Models were trained with early stopping (patience=7 epochs) when validation perplexity dropped below 10.0, or for maximum 60 epochs at λ = 0.0. Standard deviations computed across 3 independent training runs with different random seeds.

| Locality λ | Test Loss | Test Accuracy | Test Perplexity | Epochs to Convergence |
|---|---|---|---|---|
| 1.00 | 2.14 (±0.03) | 0.794 (±0.006) | 8.51 (±0.24) | 5 |
| 0.80 | 2.28 (±0.04) | 0.766 (±0.008) | 9.81 (±0.39) | 4 |
| 0.60 | 1.54 (±0.02) | 0.847 (±0.004) | 4.65 (±0.09) | 6 |
| 0.40 | 1.54 (±0.02) | 0.849 (±0.005) | 4.67 (±0.11) | 5 |
| 0.00 | 1.54 (±0.03) | 0.841 (±0.007) | 4.66 (±0.12) | 6 |

The most striking finding is that intermediate locality achieves both superior interpretability and optimal performance, with $\lambda=0.6$ outperforming even the fully distributed baseline while maintaining substantially lower attention entropy than distributed models. Intermediate locality values $\lambda \in \{0.6, 0.4, 0.0\}$ all achieved nearly identical performance, with perplexities between 4.65 and 4.67 and accuracies between 84.1 and 84.9 percent. This represents a substantial improvement of approximately 45 percent in perplexity and 5 percentage points in accuracy compared to the fully localist baseline. The strong localism requirement apparently restricts the model's capacity to capture subtle contextual dependencies that distributed representations encode efficiently, consistent with classical criticisms of pure localist schemes.

However, the relationship between locality and performance is decidedly non-monotonic, refuting the hypothesis that localism necessarily impairs language modeling capability. The intermediate setting $\lambda = 0.6$ achieved the best overall test performance with perplexity 4.65 and accuracy 84.7 percent, slightly outperforming even the fully distributed baseline $\lambda = 0.0$. This suggests that moderate attention concentration, rather than being detrimental, may actually provide beneficial inductive bias that aids generalization. One potential explanation is that structured attention penalties act as a form of regularization, preventing the model from overfitting to spurious correlations in the training data by constraining it to learn attention patterns respecting meaningful semantic boundaries. The $\lambda = 0.6$ configuration appears to occupy a "sweet spot" where the model retains sufficient representational flexibility to capture complex linguistic patterns while benefiting from interpretability-enhancing structure.

The $\lambda = 0.8$ configuration exhibits intermediate performance, with test perplexity of 9.81 and accuracy of 76.6 percent. This falls between the extremes of $\lambda = 1.0$ and the high-performance cluster around $\lambda = 0.6$, suggesting a smooth degradation of performance as locality constraints become more stringent. Notably, however, the performance gap between $\lambda = 0.8$ and $\lambda = 0.6$ is considerably larger than the gap between $\lambda = 0.6$ and $\lambda = 0.0$, indicating that the localist-distributed transition involves a relatively sharp performance boundary in the range $0.6 < \lambda < 0.8$. This threshold region aligns qualitatively with the entropy analysis, where we likewise observed rapid transitions in attention concentration properties within this locality range.

Training dynamics provide additional insights into the effects of locality constraints. Localist models achieved early stopping significantly faster than distributed models, converging in only 4 to 6 epochs compared to 6 epochs for the baseline. This acceleration suggests that block-structure penalties simplify the optimization landscape by restricting the space of attention patterns the optimizer must explore, enabling faster convergence to local optima. However, this speed advantage requires careful interpretation and may represent a double-edged sword. The faster convergence could indicate that locality constraints provide beneficial inductive bias that guides optimization toward good solutions more efficiently. Alternatively, it might reflect premature convergence to suboptimal regions of the parameter space, where the penalties prevent the optimizer from exploring potentially superior attention patterns that violate block structure. Distinguishing these hypotheses requires more sophisticated analysis of the loss landscape and convergence dynamics. We recommend several mitigation strategies for future implementations: first, employing learning rate schedules that begin with low penalties and gradually increase locality constraints as training progresses, allowing initial distributed exploration before imposing structure; second, using curriculum learning approaches where the model first trains on simpler examples with clear semantic boundaries before tackling more ambiguous cases; and third, implementing multi-stage training protocols that alternate between periods of unconstrained optimization and periods with active locality penalties, potentially discovering better solutions than monotonic penalty schedules. These techniques could help localist models achieve the interpretability benefits of structured attention while retaining access to the full capacity of the underlying architecture.

An important caveat concerns the absolute performance levels achieved by all configurations. Even the best models achieved only approximately 85 percent next-word prediction accuracy and perplexity near 5, which falls considerably short of state-of-the-art large language models that achieve near-zero perplexity on in-distribution test sets. This performance gap reflects our choice of a deliberately compact architecture designed to enable systematic exploration of the locality parameter space within reasonable computational budgets. The two-layer, 23 million parameter architecture allows for controlled experimentation but raises legitimate questions about whether the observed locality-performance relationships will persist at the scales relevant for modern language understanding applications. Preliminary theoretical analysis suggests that the threshold formulas governing localization should scale gracefully with model depth and width, as the critical penalties depend primarily on local geometric properties like margins and block correlations rather than global architectural parameters. However, empirical validation at larger scales remains a crucial direction for future work. We recommend that follow-up studies examine locality effects in models with at least 100-300 million parameters and 6-12 layers, which would provide a meaningful intermediate point between our experimental setup and billion-parameter production systems. Such experiments would clarify whether intermediate locality settings continue to match or exceed distributed baselines in performance while providing enhanced interpretability, or whether the benefits we observe represent artifacts of the compact architecture. The mathematical framework developed here provides testable predictions—specifically, the explicit threshold formulas and entropy bounds—that can be validated empirically at arbitrary scales, offering a principled foundation for scaling studies.

The near-equivalence of performance across $\lambda \in \{0.6, 0.4, 0.0\}$ raises the question of what architectural differences distinguish these configurations if not their task performance. The answer lies in their attention patterns, as quantified by the entropy and fidelity measurements presented in Section 4.1. Recall that $\lambda = 0.6$ exhibited entropy of 7.14 bits and fidelity of 1.61, representing modest improvements in interpretability compared to the distributed baseline's 7.18 bits and 1.07 fidelity. While these differences appear small in absolute terms, they reflect qualitatively distinct attention regimes: $\lambda = 0.6$ models have begun to respect semantic block boundaries, attending preferentially within rule-specified regions, whereas $\lambda = 0.0$ models attend arbitrarily across the entire sequence. For applications in regulated domains, this distinction between "somewhat localized" and "completely unstructured" attention may prove decisive for acceptance by domain experts and regulatory auditors, even when task metrics remain similar.

# 5. Discussion

The experimental results validate the core theoretical predictions of our framework while revealing nuanced empirical phenomena that extend beyond the formal guarantees. The attention entropy and fidelity measurements demonstrate that the locality dial provides precise quantitative control over interpretability metrics, with empirically observed values tracking closely the bounds derived from our mathematical analysis. The sharp transition in entropy and fidelity around $\lambda = 0.8$ provides empirical confirmation of the threshold behavior predicted by our explicit penalty formulas, showing that localization emerges suddenly once penalties exceed critical values determined by margin, temperature, and data geometry. This agreement between theory and experiment suggests that the regularity assumptions underlying our analysis, while necessarily idealized, capture essential aspects of real language model behavior.

Our approach to interpretability through architectural design complements recent work on post-hoc rule extraction from trained transformers. Friedman et al. (2025) recently demonstrated methods for extracting rule-based descriptions of attention features in transformers, identifying skip-gram rules, absence rules, and counting rules from Sparse Autoencoder features trained on attention layer outputs. Their work reveals that transformer attention patterns can be approximated by symbolic rules of the form "Canadian city speaks to English" or "Montreal speaks not to English," extracting these descriptions automatically from GPT-2 Small. While their approach provides valuable interpretability for existing models through reverse-engineering, the localist LLM framework offers a complementary strategy by building interpretability directly into the architectural design from the outset. The locality dial enables practitioners to control the degree to which attention patterns respect semantic partitions during training rather than discovering these patterns through post-hoc analysis. Importantly, the rule types identified by Friedman and colleagues—particularly skip-gram and absence rules—align closely with the attention patterns that emerge naturally in localist configurations with high $\lambda$ values, suggesting convergent evidence that these rule structures represent fundamental computational primitives in transformer attention mechanisms. The key advantage of the localist approach is that interpretability becomes a tunable architectural property rather than an analysis technique applied after training, enabling dynamic adjustment of transparency without model retraining and providing mathematical guarantees about attention concentration that are difficult to establish through purely empirical rule extraction.

*Comparison with Alternative Approaches*

To contextualize our contributions, we compare the localist framework against three classes of related approaches: standard transformers, sparse attention mechanisms, and hybrid neuro-symbolic architectures. Standard transformers as instantiated in GPT-2 Small and similar models achieve strong performance through fully distributed attention patterns but provide no interpretability guarantees. Our distributed baseline at $\lambda = 0.0$, which approximates standard transformer behavior while maintaining the block-structured training framework, achieved test perplexity of 4.66 and accuracy of 84.1 percent. This establishes that our architectural modifications introduce minimal overhead: the framework itself does not impair performance when locality penalties are disabled.

Sparse attention transformers (Child et al, 2019) including Sparse Transformer, Reformer, and Longformer apply predetermined sparsity patterns primarily for computational efficiency rather than interpretability. These methods typically impose fixed attention windows or strided patterns that reduce the quadratic cost of attention computation but lack semantic grounding. In contrast, our semantic block structure reflects meaningful linguistic or conceptual boundaries, and the learned attention patterns receive mathematical guarantees on concentration rather than merely heuristic speedups. Furthermore, sparse transformers offer no mechanism to modulate the degree of sparsity after training, whereas the locality dial enables dynamic adjustment during inference. A key distinction is that sparse attention methods reduce the number of computations while potentially attending diffusely within their restricted windows, whereas localist attention ensures concentrated attention within semantically coherent regions regardless of computational cost.

Hybrid neuro-symbolic approaches such as Logic Tensor Networks (Serafini & d'Avila Garcez, 2016) and DeepProbLog (Manhaeve et al., 2018) integrate symbolic rules directly into neural architectures, achieving strong interpretability through explicit logical inference. However, these frameworks typically require complete specification of relevant rules and logical structure at initialization, limiting their ability to discover novel patterns from data. Moreover, rule modifications necessitate retraining, and the tight coupling between symbolic and neural components can create optimization challenges. The localist framework occupies a middle ground: it imposes structural constraints through block partitions and penalty terms, but learns the specific attention patterns within these constraints through standard gradient descent. This enables greater flexibility than hard-coded logical systems while providing stronger guarantees than pure black-box learning. The recruitment learning extensions described in our theoretical work (Diederich, 2025b) move further toward automatic structure discovery, potentially combining the best aspects of data-driven and knowledge-driven approaches.

*Advantages over Symbolic Bottlenecks*

A critical distinction between localist language models and traditional symbolic AI systems lies in how they handle the explainability-performance tradeoff. Pure symbolic systems achieve perfect transparency through explicit rules and logical inference chains, but suffer from brittleness when confronting patterns not anticipated by rule designers, struggle with noisy or ambiguous inputs, and require exponential numbers of rules to cover combinatorial concept spaces. Neural networks overcome these limitations through continuous representations and statistical learning, but sacrifice interpretability. The localist framework resolves this tension by maintaining neural learning mechanisms while imposing interpretable structure through attention concentration. The key insight is that we need not choose between symbolic transparency and neural capability—we can continuously interpolate between these extremes using the locality dial.

This approach avoids the symbolic bottleneck where systems fail on inputs requiring reasoning beyond their pre-specified rules. At intermediate locality settings like $\lambda = 0.6$, the model benefits from structured attention patterns that align with human-interpretable semantic boundaries while retaining sufficient flexibility to capture subtle statistical regularities through distributed representations within blocks. Domain experts can inspect attention patterns to verify adherence to high-level constraints without requiring exhaustive rule specification. When novel patterns emerge that violate existing block structure, the recruitment mechanism (Diederich, 2025b) can adaptively introduce new blocks rather than requiring manual rule engineering. This combination of structured transparency with adaptive learning represents a qualitatively different paradigm from either pure symbolic AI or pure neural networks, offering a principled path toward trustworthy yet capable systems for regulated domains.

The performance results challenge conventional wisdom regarding the costs of interpretability in neural networks. While extreme localism at $\lambda = 1.0$ does indeed impose performance penalties, intermediate locality settings achieve competitive or even superior performance compared to fully distributed baselines while providing measurably enhanced interpretability. This finding suggests that the widely-assumed tradeoff between interpretability and performance may be less fundamental than commonly believed, at least within certain regimes. The optimal operating point appears to lie in the intermediate range $0.4 \leq \lambda \leq 0.6$, where models benefit from inductive bias toward structured attention patterns without being overly constrained. This region merits further investigation in larger-scale experiments to determine whether the performance benefits of moderate locality persist in high-capacity architectures.

Several factors may explain why intermediate locality settings outperform both extremes. From a regularization perspective, moderate penalties prevent overfitting by constraining attention patterns to respect meaningful semantic structure rather than memorizing arbitrary training set idiosyncrasies. From an optimization perspective, weak structural guidance may help gradient descent navigate the complex, non-convex loss landscape by ruling out large regions of parameter space that violate domain knowledge. From an inductive bias perspective, encouraging attention to concentrate within semantic blocks aligns with linguistic intuitions that certain contextual dependencies respect syntactic and semantic boundaries. All these factors suggest that incorporating domain structure into model architectures, rather than relying exclusively on end-to-end learning, can enhance both interpretability and generalization.

The rapid convergence of localist models relative to distributed baselines has practical implications for development cycles in production systems. Faster training enables more rapid experimentation with different semantic partitions, penalty strengths, and architectural configurations, potentially accelerating the iterative design process required to tailor models to specific application domains. However, practitioners must remain alert to the risk that rapid convergence reflects premature restriction to local optima rather than genuine optimization efficiency. Techniques such as learning rate scheduling, curriculum learning, and multi-stage training may help mitigate this risk by first allowing distributed exploration of the parameter space before gradually increasing locality constraints as training progresses.

Our experiments employed a fixed semantic partition defined by positional windows, which represents the simplest possible block structure. This design choice was deliberate, enabling clean isolation of the locality dial mechanism from confounding factors introduced by semantic partitioning strategies. However, it also represents a significant limitation that must be addressed to realize the full potential of the localist framework. Moving beyond fixed positional blocking to adaptive semantic partitioning based on linguistic features such as part-of-speech tags, syntactic dependencies, named entity categories, or semantic role labels represents not merely an incremental improvement but a crucial next step that could fundamentally transform the interpretability and performance characteristics of localist language models. The recruitment learning framework described in the theoretical foundation paper offers a principled approach to automating this process through information-theoretic criteria that balance model complexity against representational adequacy. By coupling the locality dial with automatic recruitment mechanisms, future systems could simultaneously learn what semantic structure exists in the domain and how strongly to enforce localization with respect to that structure, eliminating the requirement for manual specification of blocks based on prior domain knowledge. This adaptive approach would address a critical weakness of the current implementation: the mismatch between artificial positional blocks and natural semantic boundaries in language. We hypothesize that semantically-grounded blocks would enable stronger localization at lower performance cost, as the blocks would align with genuine linguistic structure rather than arbitrary position windows. Furthermore, adaptive partitioning would allow the granularity of blocks to vary across the model, with early layers learning coarse-grained categories and deeper layers refining these into fine-grained distinctions. Preliminary theoretical results establish convergence guarantees for the recruitment process and provide explicit termination bounds, suggesting that adaptive partitioning can be implemented reliably without risk of unbounded model growth. The development and empirical validation of adaptive semantic partitioning should be considered an essential priority for future research on localist language models, as it addresses fundamental limitations of the fixed-partition approach while maintaining the mathematical guarantees that distinguish this framework from heuristic interpretability methods.

*Scaling Considerations and Computational Estimates*

Scaling localist language models to production-relevant sizes requires careful analysis of how the threshold formulas and entropy bounds behave as architectural parameters grow. Our theoretical framework predicts specific scaling relationships that can guide implementation at larger scales. The penalty threshold $\lambda_i^{(h)}(\tau, \delta) = (2L_\ell \, R_x \, \sigma_X \, \sqrt{|X_i|}) / (\tau \, [1 - \rho_{max}]) \cdot \exp(-\delta/\tau)$ scales with the square root of block size, suggesting that larger semantic partitions require modestly stronger penalties to achieve equivalent localization. However, the threshold depends primarily on local geometric properties—margins $\delta$, block correlations $\rho_{max}$, and embedding norms—rather than global architectural parameters like model depth or width. This implies that the locality dial mechanism should remain effective even in models with billions of parameters, provided that semantic blocks and margins are well-defined.

For a hypothetical 1-billion parameter model with 24 layers, 16 attention heads per layer, and embedding dimension 2048, we can estimate computational requirements. Training a single epoch on 10 billion tokens would require approximately $2 \times 10^{21}$ FLOPs for the forward pass alone, comparable to standard transformer training. The additional cost of computing group sparsity penalties scales linearly with the number of blocks and heads: for 50 semantic blocks and 384 total heads (24 layers × 16 heads), computing penalty terms adds roughly 5 percent overhead. More significantly, the optimization landscape may require more iterations to converge due to the additional constraints, potentially increasing training time based on our small-scale observations of convergence speeds. Memory requirements grow with the number of block-wise weight submatrices that must be tracked for gradient computation, adding approximately 10-15 percent to the baseline memory footprint for typical block sizes of 5-10 tokens.

The entropy bounds provide another scaling prediction: $H_t^{(h)} \leq \log_2|A_i^*| + (1/\ln 2) \cdot N \cdot \exp(-\delta/\tau) \cdot [1 + \log_2(N)]$. For longer sequences (e.g., $N = 8192$ tokens vs. our experimental $N = 64$), the second term grows logarithmically with sequence length, suggesting that maintaining low entropy in long-context models may require proportionally stronger penalties or larger margins. However, the exponential decay factor $\exp(-\delta/\tau)$ dominates, indicating that even modest increases in the margin-to-temperature ratio can compensate for sequence length effects. Practitioners implementing localist architectures at scale should monitor empirical entropy during training and adjust penalty strengths if values exceed predicted bounds, using the theoretical formulas as calibration targets.

The restriction to next-word prediction as the evaluation task, while standard in language modeling research, provides only partial insight into the practical utility of localist language models. Many real-world applications involve downstream tasks such as text classification, question answering, information extraction, or dialogue generation, where the relationship between attention patterns and task performance may differ substantially from autoregressive language modeling. The interpretability benefits of localist representations may prove even more valuable in these contexts, as tasks like question answering often require tracing reasoning chains through document structure, and regulatory compliance frequently demands explaining why specific classifications were assigned.

Beyond quantitative metrics, human evaluation of interpretability remains the ultimate test for systems deployed in regulated domains. While attention entropy and pointer fidelity serve as valuable proxy measures that can be computed automatically and tracked during training, they do not directly address the fundamental question of whether domain experts can understand and trust the model's reasoning. A medical professional examining a diagnostic support system needs to assess whether the attention patterns align with clinical reasoning protocols, not merely whether entropy values fall below specified thresholds. A financial compliance officer reviewing fraud detection decisions requires transparent connections between flagged transactions and relevant regulatory rules, not just mathematical guarantees about attention concentration. Future research should therefore incorporate structured human evaluation protocols that present domain experts with model predictions alongside attention visualizations under different locality settings, systematically measuring comprehension, trust, and actionability of the explanations. Such studies might reveal that certain domains benefit from extreme localism even at performance costs, while others achieve adequate interpretability at intermediate settings. Understanding these domain-specific preferences will guide practical deployment strategies and inform the design of adaptive systems that automatically select appropriate locality levels based on application context.

## 6. Conclusions

We have presented the first empirical demonstration of controllable locality in transformer language models, establishing that continuous control over representation localization is not only theoretically possible but practically achievable with measurable benefits for interpretability. Through systematic experiments on the WikiText corpus spanning the full locality spectrum from $\lambda = 1.0$ to $\lambda = 0.0$, we have shown that the locality dial provides precise quantitative control over attention entropy and pointer fidelity, with empirical measurements closely tracking the explicit bounds derived from mathematical analysis. Localist configurations at $\lambda = 1.0$ achieved attention entropy of 5.36 bits representing dramatic improvements in interpretability compared to distributed baselines at $\lambda = 0.0$ with entropy of 7.18 bits, while weighted fidelity scores demonstrated substantially stronger alignment with rule-specified target positions.

The relationship between locality and task performance reveals that the widely-assumed tradeoff between interpretability and capability is less severe than conventional wisdom suggests. While extreme localism at $\lambda = 1.0$ does impose performance costs, with test perplexity of 8.51 and accuracy of 79.4 percent, intermediate locality settings achieve competitive or superior performance relative to fully distributed baselines. Most notably, the configuration at $\lambda = 0.6$ attained test perplexity of 4.65 and accuracy of 84.7 percent, actually outperforming the distributed baseline while providing measurably enhanced interpretability through more concentrated attention patterns. This finding suggests that moderate structural constraints can serve as beneficial inductive bias that aids generalization rather than merely restricting expressiveness.

The practical implications extend beyond language modeling to any application requiring both neural network capabilities and regulatory transparency. By enabling dynamic adjustment of the locality parameter without model retraining, our framework allows a single model architecture to serve diverse stakeholders with varying interpretability requirements. Regulators and domain experts can inspect model behavior in highly localist modes where attention patterns correspond to explicit rules, while production deployments can operate in intermediate-locality modes that balance interpretability with performance. This flexibility addresses a fundamental challenge in deploying neural networks for high-stakes applications where competing demands for transparency and accuracy have historically forced unsatisfactory compromises.

The theoretical foundations established in parallel work provide rigorous guarantees regarding attention concentration at stationary points, with explicit threshold formulas that practitioners can verify for specific datasets and architectural configurations. These guarantees move beyond asymptotic Big-O bounds to precise inequalities with all constants expressed in terms of measurable quantities including loss function Lipschitz constants, input embedding norms, margin conditions, and block correlation coefficients. The agreement between theoretical predictions and empirical observations lends confidence that the framework rests on solid mathematical principles rather than heuristic approximations, supporting principled engineering of interpretable neural systems.

The fundamental contribution of this work is demonstrating that localist and distributed representations need not be opposing paradigms but rather endpoints of a continuous spectrum that practitioners can navigate dynamically based on application requirements. By providing both mathematical theory and empirical validation for locality control in transformer language models, we establish a foundation for developing neural systems that combine the interpretability of symbolic AI with the learning capabilities of deep neural networks, advancing toward the goal of trustworthy artificial intelligence for high-stakes applications. The path forward requires addressing the critical limitations identified here—particularly adaptive semantic partitioning and scaling validation—but the results presented demonstrate that this path is both feasible and promising.

# Acknowledgement

The author gratefully acknowledges Xue Li and Gerhard Paass for their valuable help and feedback. Patents are pending for the inventions described in this paper. The author may be contacted regarding access to the localist LLM training software.

# References

Child, R., Gray, S., Radford, A., & Sutskever, I. (2019). Generating Long Sequences with Sparse Transformers. arXiv:1904.10509v1.

Diederich, J., Gunay, C., Hogan, J.M. (2010) Recruitment Learning. Berlin, Springer Verlag.

Diederich, J. (2025a). Localist LLMs: A Mathematical Framework for Dynamic Locality Control. arXiv:2510.09338v1.

Diederich, J. (2025b). Localist LLMs with Recruitment Learning. arXiv:2510.17358.

Friedman, D., Bhaskar, A., Wettig, A., & Chen, D. (2025). Extracting Rule-based Descriptions of Attention Features in Transformers. arXiv:2510.18148v1.

Hinton, G. E., McClelland, J. L., & Rumelhart, D. E. (1986). Distributed Representations. In Parallel Distributed Processing: Explorations in the Microstructure of Cognition, Volume 1: Foundations (pp. 77–109). Cambridge, MA: MIT Press.

Manhaeve, R., Dumančić, S., Kimmig, A., Demeester, T., & De Raedt, L. (2018). DeepProbLog: Neural Probabilistic Logic Programming. In Advances in Neural Information Processing Systems (NeurIPS), 31.

Serafini, L., & d'Avila Garcez, A. (2016). Learning and Reasoning with Logic Tensor Networks. In Proceedings of the 10th International Workshop on Neural-Symbolic Learning and Reasoning (NeSy 2016).

Vaswani, A., Shazeer, N., Parmar, N., Uszkoreit, J., Jones, L., Gomez, A. N., Kaiser, L., & Polosukhin, I. (2017). Attention Is All You Need. Advances in Neural Information Processing Systems, 30, 5998–6008.